%% file: main.tex
\documentclass[10pt,twocolumn,letterpaper]{article}

\usepackage[pagenumbers]{cvpr} 

\input{preamble}

\definecolor{cvprblue}{rgb}{0.21,0.49,0.74}
\usepackage[pagebackref,breaklinks,colorlinks,allcolors=cvprblue]{hyperref}

\title{ A Neurosymbolic Framework for Interpretable Cognitive Attack Detection in Augmented Reality}

\author{
Rongqian Chen$^1$ \quad Allison Andreyev $^1$ \quad Yanming Xiu$^2$ \quad Joshua Chilukuri$^2$ \quad Shunav Sen$^2$ \\
\quad Mahdi Imani$^4$ \quad Bin Li$^3$ \quad Maria Gorlatova$^2$ \quad Gang Tan$^3$ \quad Tian Lan$^1$\\[6pt]
$^1$George Washington University \qquad $^2$Duke University \\
$^3$Pennsylvania State University \qquad $^4$Northeastern University \\
}

\begin{document}
\maketitle
\input{sec/0_abstract}    
\input{sec/1_intro}

\input{sec/2_relatedWork}

\input{sec/3_methodology}
\input{sec/4_experiment}
\input{sec/5_conclusion}
{
    \small
    \bibliographystyle{ieeenat_fullname}
    \bibliography{main}
}

\input{sec/6_appendix}

\end{document}

%% file: preamble.tex
\usepackage{multirow}
\usepackage{array}
\usepackage[table]{xcolor}
\definecolor{lightgray}{gray}{0.9}

%% file: sec/0_abstract.tex
\begin{abstract}

Augmented Reality (AR) enriches human perception by overlaying virtual elements onto the physical world. However, this tight coupling between virtual and real content makes AR vulnerable to cognitive attacks: manipulations that distort users’ semantic understanding of the environment. Existing detection methods largely focus on visual inconsistencies at the pixel or image level, offering limited semantic reasoning or interpretability. To address these limitations, we introduce CADAR, a neuro-symbolic framework for cognitive attack detection in AR that integrates neural and symbolic reasoning. CADAR fuses multimodal vision–language representations from pre-trained models into a perception graph that captures objects, relations, and temporal contextual salience. Building on this structure, a particle-filter-based statistical reasoning module infers anomalies in semantic dynamics to reveal cognitive attacks. This combination provides both the adaptability of modern vision–language models and the interpretability of probabilistic symbolic reasoning. Preliminary experiments on an AR cognitive-attack dataset demonstrate consistent advantages over existing approaches, highlighting the potential of neuro-symbolic methods for robust and interpretable AR security.
\end{abstract}

%% file: sec/1_intro.tex
\section{Introduction}
\label{sec:intro}

\begin{figure}[t]
    \centering
    \includegraphics[width=3.3in]{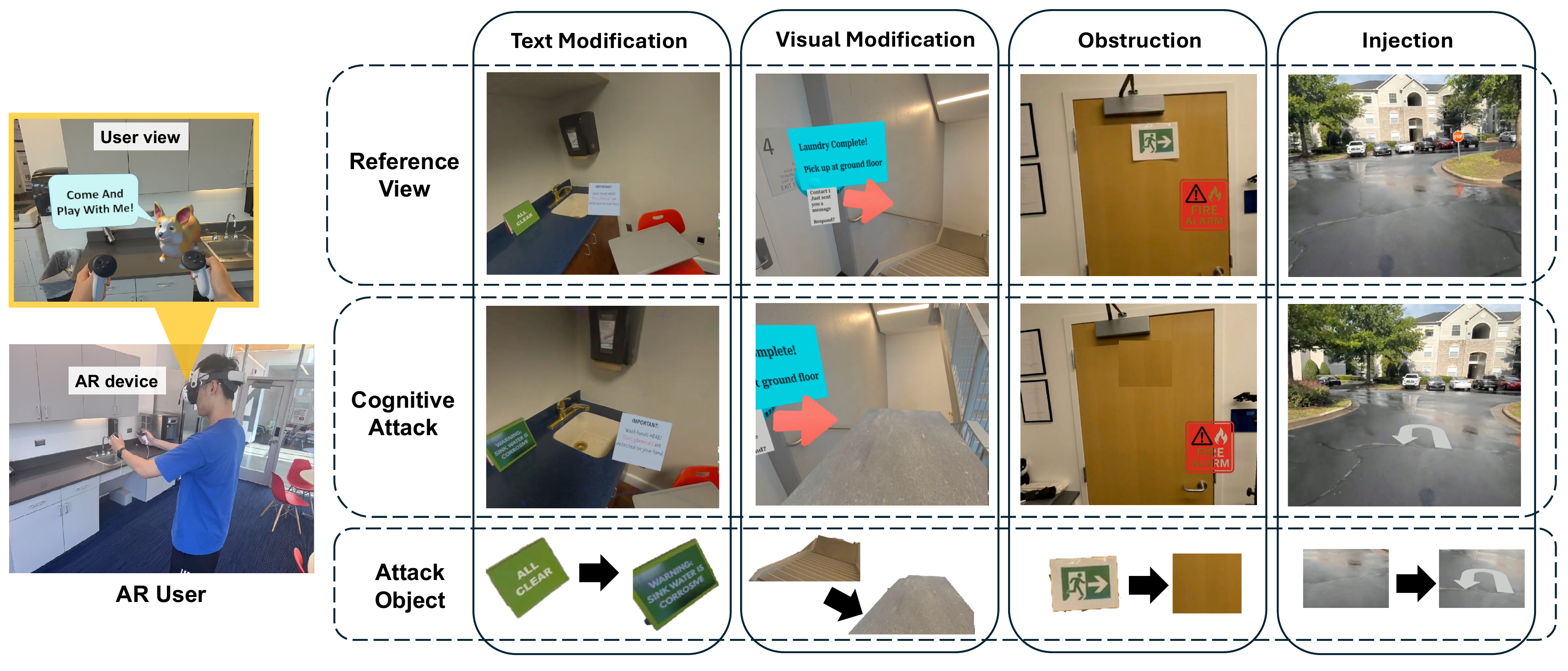}
    \vspace{-0.25in}
    \caption{Illustrative AR/MR cognitive attacks. Top row: original scenes; bottom row: manipulated scenes. (a) Text Modification Attack. (b) Visual Modification attack. (c) Obstruction Attack. (d) Injection Attack.}
     \vspace{-0.2in}
    \label{cognitive attacks}
\end{figure}

Augmented Reality (AR) and Mixed Reality (MR) technologies increasingly overlay virtual content onto the physical world, reshaping how users perceive and interact with their surroundings~\cite{zheng2023review}. This tight integration creates new vulnerabilities, as users tend to trust overlaid visual and spatial cues. Adversaries can exploit this trust to launch cognitive attacks, where malicious virtual content is injected or altered to mislead, distract, or manipulate users without their awareness~\cite{10924575,cheng2023exploring, teymourian2025sok,xiu2025detectingvisualinformationmanipulation}. Unlike traditional cybersecurity threats that target system integrity, cognitive attacks operate at the perceptual level, subtly manipulating users’ semantic understanding of a scene (examples in Fig.\ref{cognitive attacks}).


Prior work on detecting anomaly and attacks often employs computer vision techniques to identify visual changes or adversarial distortions~\cite{qiu2019review, gu2024anomalygpt,qu2023towards, wang2022objectformer, duan2022confusing, li2024survey, acharya2022detecting}. 
However, these methods rely on pixel/image processing, lacking the ability to model the human perception process that not only involves prior knowledge and contextual information, but also selectively focuses on aspects of the visual field for semantical interpretation. As a result, these methods may struggle to capture smart cognitive attacks manipulating only a small number of pixels and also may suffer from false positives due to benign visual perturbations that do not affect users' semantic interpretation~\cite{li2024survey}. 
More recent studies propose pre-trained vision–language models (VLMs)~\cite{xiu2025viddar, roy2025zero} can achieve semantic-level detection by processing visual features and incorporating human-like knowledge. But the reasoning of pre-trained VLMs is a black-box approach offering limited interpretability. It is difficult to leverage domain-specific knowledge or model scene/context evolutions or temporal correlations in cognitive attack detection. 

In this paper, we consider cognitive attacks that aim to manipulate users' semantic perception in AR/MR.  We propose \emph{CADAR} (Cognitive Attack Detection in Augmented Reality), a neuro-symbolic approach encompassing two key components: (i) a symbolic perception-graph model, inspired by knowledge and scene graphs~\cite{hogan2021knowledge,chang2021comprehensive,yun2025missiongnn,shah2022lmnavrobotic}. It extracted using pre-trained neural VLM models to represent scene/context evolution at the semantic-level, and (ii) a particle-filter model, which is commonly used in non-linear and stochastic systems. It applies a sequential Monte Carlo method to the perception graph for reasoning and detecting cognitive attacks, stabilizes inherently noisy or inconsistent VLM outputs, and mitigates errors or missed detections through probabilistic fusion. It also offers granular, flexible temporal detection, using reference sequences, frame-wise inference localizes attacks per frame, to produce a variable number of labels when needed.
Thus, the \emph{CADAR} approach inherits the adaptability of the pre-trained VLM (neural part), as well as the interpretability and reasoning rigor of particle filtering (symbolic part). It semantically interprets and models the sequential scene changes for robust cognitive attack detection in AR/MR, overcoming existing work's limitations.


We summarize our contributions as follows:
\begin{itemize}
    \item \textbf{Neuro-symbolic framework.} We introduce a hybrid architecture that converts VLM outputs into symbolic perception graphs for explicit and interpretable reasoning in AR cognitive attack detection.
    
    \item \textbf{Perception graph schema.} We design a spatio-temporal graph representation that integrates multimodal information from image/text encoders and detection modules to capture scene evolution.

    \item \textbf{Particle-filter reasoning.} We apply particle filtering to symbolic perception graphs, improving robustness to noisy VLM outputs and enabling principled statistical anomaly detection.

    \item \textbf{CADAR-50K dataset.} We build the first public AR cognitive attack dataset and demonstrate that \emph{CADAR} achieves 74.6\% accuracy, significantly outperforming all baseline methods.
\end{itemize}

The rest of the paper is organized as follows: Section 2 reviews related work; Section 3 presents our method; Section 4 reports experimental results; and Section 5 concludes the paper.

%% file: sec/2_relatedWork.tex
\section{Related Work}

\paragraph{Cognitive Attacks in AR/MR.} 

One well-studied category of visual attacks is adversarial attacks, which introduce imperceptible yet malicious perturbations to input data to mislead machine learning models~\cite{zhang2024adversarial,mumcu2024multimodal}. In contrast, cognitive attacks target perception-level semantics and aim to alter a human user’s understanding of a scene~\cite{cheng2023exploring,xiu2025detectingvisualinformationmanipulation}. Augmented Reality (AR) systems, which overlay virtual content onto the physical world to enhance perception and decision-making, are particularly vulnerable to such attacks. However, research on cognitive attacks in AR remains limited. Existing methods often rely on access to the original, unaltered scene for comparison or employ pre-trained VLMs with limited interpretability.

\vspace{-3mm}
\paragraph{Neuro-symbolic Methods.} 
Neuro-symbolic approaches couple deep neural networks with symbolic reasoning--i.e., reasoning over explicitly represented knowledge in formal languages, to capitalize on the complementary strengths of subsymbolic pattern recognition and symbolic compositionality~\cite{bhuyan2024neuro,nawaz2025review}. Within knowledge-graph research, most neuro-symbolic work addresses link prediction or graph completion, where the hybrid models offer interpretable rules, guided learning, and more transparent decision pathways~\cite{delong2024neurosymbolic}. Yet, like standard supervised techniques, many of these solutions still demand sizable labeled datasets. In contrast, our framework introduces a neuro-symbolic reasoning module that combines explicitly encoded rules with statistics, achieving high interpretability while remaining data-efficient.
\vspace{-3mm}
\paragraph{Particle Filter and Statistical Reasoning.}
Particle filter is a sequential Monte Carlo algorithms for recursive Bayesian estimation~\cite{chen2003bayesian}, it propagates a set of weighted samples to approximate the posterior state distribution, outperforming classical Kalman filters in strongly non-linear or non-Gaussian settings~\cite{kunsch2013particle,5985520}. They are widely used in surveillance, robotics, and navigation, 
leveraging visual, geometric, and motion cues such as color, texture, and shape \cite{awal2023particle}. In \emph{CADAR}, we model graph nodes and edges as stochastic particles, enabling robust reasoning under noisy or inconsistent VLM outputs.

%% file: sec/3_methodology.tex
\section{Methodology}

\begin{figure}[t]
    \centering
    \includegraphics[width=3.4in]{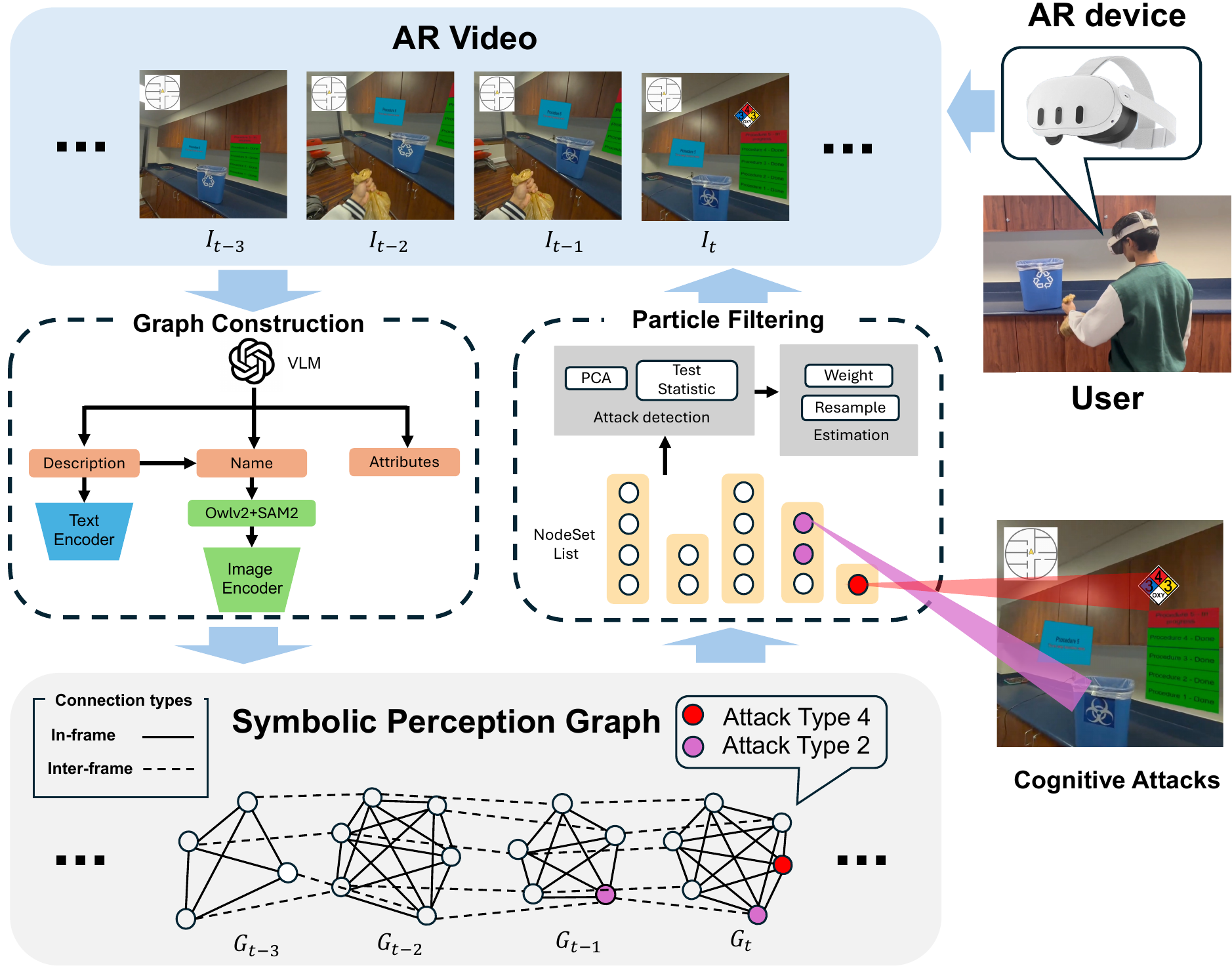}
    \vspace{-0.2in}
    \caption{A system overview of \emph{CADAR}. Sequential video frames are first transformed into spatial-temporal symbolic perception graphs. The attack-detection module then analyzes these graphs to identify, classify, and localize adversarial attacks, such as the visual modification and injection attacks shown in the examples.}
        \vspace{-1mm}
    \label{fig:system framework}
\end{figure}

In this section, we define the four types of cognitive attacks in AR scenes and introduce our neuro-symbolic framework for detecting such attacks. As illustrated in Fig.~\ref{fig:system framework}, the framework consists of two main components: the perception graph model and the particle filter-based detection module. The perception graph encodes each frame as a 3D spatiotemporal structure, where nodes represent objects and edges capture their relationships across time (Fig.~\ref{fig:perception graph fig}). These structured graphs are then passed into the attack detection algorithm (Fig.~\ref{fig: attack detection algorithm}), which reasons over the particle filter to identify cognitive attacks.

\subsection{Cognitive Attacks}
\label{sec:attacks}
We investigate four representative cognitive attack types that deliberately compromise the semantic integrity of AR scenes (Fig.\ref{cognitive attacks}). The text-modification attack manipulates on-scene text—such as changing “\textsc{No Parking}” to “\textsc{Free Parking}”—to reverse intended meanings or inject false instructions, thereby deceiving both human users and perception systems. The visual-modification attack distorts object appearance or placement—for example, turning a green light to red or relocating a stop sign—leading to recognition errors and misinformed user decisions. The obstruction attack targets critical information by occluding or deleting essential cues like exit signs, disrupting safety awareness and expected graph relations. Finally, the injection attack introduces fictitious elements, such as fake hazard symbols or virtual labels, that embed misleading cues, divert user attention, and corrupt subsequent reasoning processes.

\subsection{Perception Graph Model}

\begin{figure*}[t]
    \centering
    \includegraphics[width=6.8in]{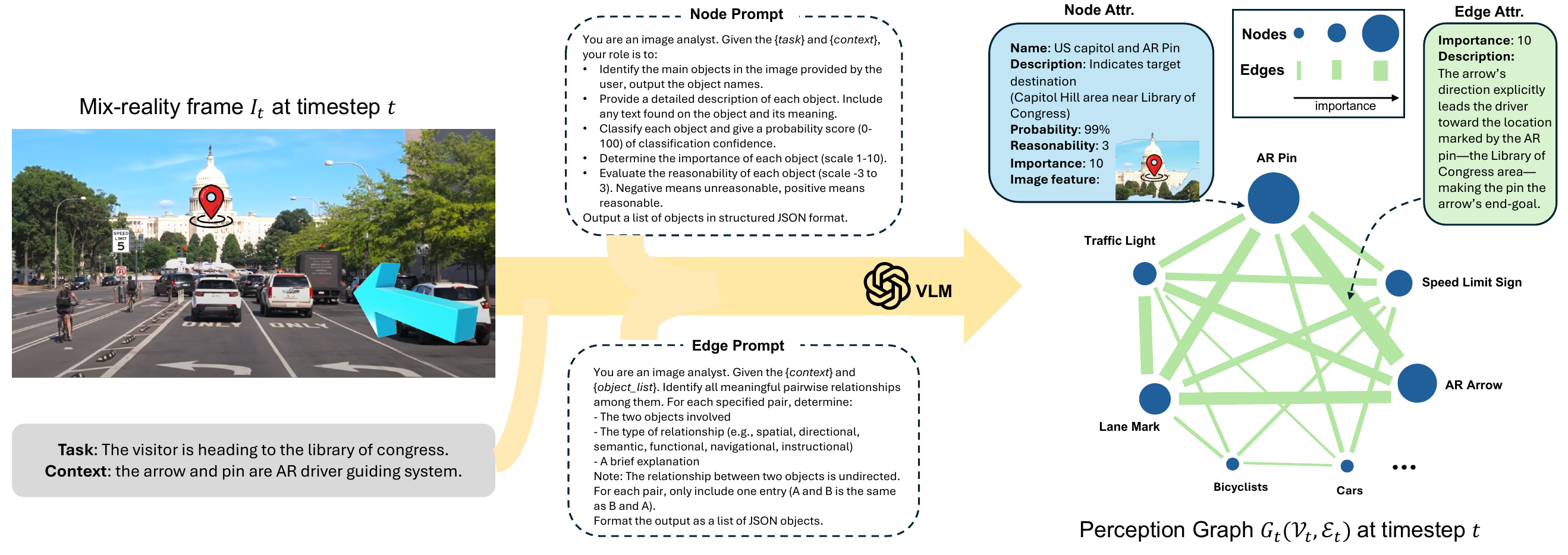}\vspace{-0.1in}
    \caption{Symbolic perception graph generation: given a video frame and its contextual description, VLMs generate the corresponding perception graph at time step t.}
     \vspace{-0.1in}
    \label{fig:perception graph fig}
    \vspace{-1mm}
\end{figure*}

The perception graph model is a symbolic representation of perception, simulating how humans perceive, interpret, and prioritize information. It collects, structures, and stores scene data that enables formal reasoning later. The graph leverages pre-trained VLMs to approximate human interpretation of sensory input. Given AR imagery and task context, the model fuses multimodal signals into a compact graph and embeds their semantics in a latent perceptual space.

A perception graph consists of nodes and edges. Objects are represented as nodes, and object relationships are represented as edges. Given the context information and prompt (shown in the Fig.\ref{fig:perception graph fig}), the VLM would extract the related objects with their relationships in the scene. At timestep $t$, the Perception Graph $G_t$ is generated as follows.

\subsubsection{Graph Construction.}
At discrete timestep $t\!\in\!\{1,\dots,T\}$ we ingest  
\vspace{-2mm}

\begin{equation}
    I_t \in \mathbb{R}^{H\times W\times 3},\qquad 
   c_t \in \mathcal{C},
\end{equation}

\noindent where $I_t$ is the AR video frame and $c_t$ encodes task context. A pre-trained vision-language module $\mathcal{F}_{\text{VLM}}$ lifts the frame into a spatio-temporal perception graph
\begin{equation}
    G_t = (\mathcal{V}_t,\mathcal{E}_t) 
   = \mathcal{F}_{\text{VLM}}\!\bigl(I_t,\,c_t\bigr),
\end{equation}

\noindent with nodes $\mathcal{V}_t$ and edges $\mathcal{E}_t$:
\vspace{-2mm}
\begin{equation}
    \mathcal{V}_t = \{v_i^t\}_{i=1}^{N_t}, 
   \qquad
   \mathcal{E}_t = \{\,e_{ij}^t \mid v_i^t,v_j^t\in\mathcal{V}_t\}.
\end{equation}
   \vspace{-3mm}
   
For reasoning that needs the \emph{entire} observation history
(e.g. long-range causal inference) we define GraphSet
\vspace{-3mm}
\begin{equation}
\hat{G}_t = (\mathcal{V}_{1:t},\, \mathcal{E}_{1:t}),\quad
\mathcal{V}_{1:t}=\bigcup_{s=1}^{t}\mathcal{V}_s,\quad
\mathcal{E}_{1:t}=\bigcup_{s=1}^{t}\mathcal{E}_s.
\end{equation}
\vspace{-3mm}

Thus, the GraphSet $\hat{G}_t$ is the union of all spatial and temporal connections that stitch object instances across time steps $1{:}t$. Each node still carries its original timestamp to preserve the temporal ordering.

Each node in the perception graph is represented by several attributes, including its ID, attack type, and timestep, along with a semantic description, visual features such as segmentation masks, a recognition probability score, contextual importance based on task relevance, and a reasonability score ranging from $-3$ to $+3$. These attributes are encoded using pre-trained vision-language and segmentation models to capture multimodal object representations. Each edge connects two nodes and includes similar parameters (node IDs, attack type, and timestep), along with a textual description capturing semantic, functional, navigational, or instructional relationships. Edge importance is derived from the contextual salience of the connected nodes.

\subsubsection{NodeSet and EdgeSet.}



After defining the temporal graph structure $\hat{G}_t$, which links perception graphs across time, we introduce the NodeSet and EdgeSet to represent the temporally connected nodes and edges, respectively.

At each time step $t$, the NodeSet for object $i$ is denoted as $\mathcal{N}_{i,t}$, and the collection of all such NodeSets is represented as the NodeSet List $\mathcal{N}_t$. 

\begin{equation}
\begin{aligned}
\mathcal{N}_{i,t}
&= \{\, v_i^k \mid 1\le k\le t \,\}, \qquad i\in\mathcal{I}_t,\\
\mathcal{N}_t
&= \{\, \mathcal{N}_{i,t} \mid i\in\mathcal{I}_t \,\}.
\end{aligned}
\end{equation}

\noindent where
\(
   \mathcal{I}_t = \bigcup_{s=1}^{t}\{\,i \mid v_i^s\in\mathcal{V}_s\}
\)
is the set of all object IDs observed so far.
      
Similarly, EdgeSet and EdgeSet List is defined as:
\begin{equation}
\begin{aligned}
E_{ij,t}
&= \{\, e_{ij}^{k} \mid 1\le k\le t \,\}, \qquad i,j\in\mathcal{I}_t,\\
E_t
&= \{\, E_{ij,t} \mid i,j\in\mathcal{I}_t \,\}.
\end{aligned}
\end{equation}
\vspace{-2mm}

Thus the GraphSet $\hat{G}_t$ can also represented by NodeSet and EdgeSet:
\begin{equation}
    \hat{G}_t
   =\bigl(
        \mathcal{V}_{1:t},\,
        \mathcal{E}_{1:t}
     \bigr), \hspace{1mm}\mathcal{V}_{1:t}=\!\!\bigcup_{i\in\mathcal{I}_t}\!\!\mathcal{N}_{i,t}, \hspace{1mm}
    E_{1:t}=\!\!\bigcup_{i,j\in\mathcal{I}_t}\!\!\mathcal{E}_{ij,t}.
\end{equation}

\subsection{Particle Filter Model}

\begin{figure*}[t]
    \centering
    \includegraphics[width=6in]{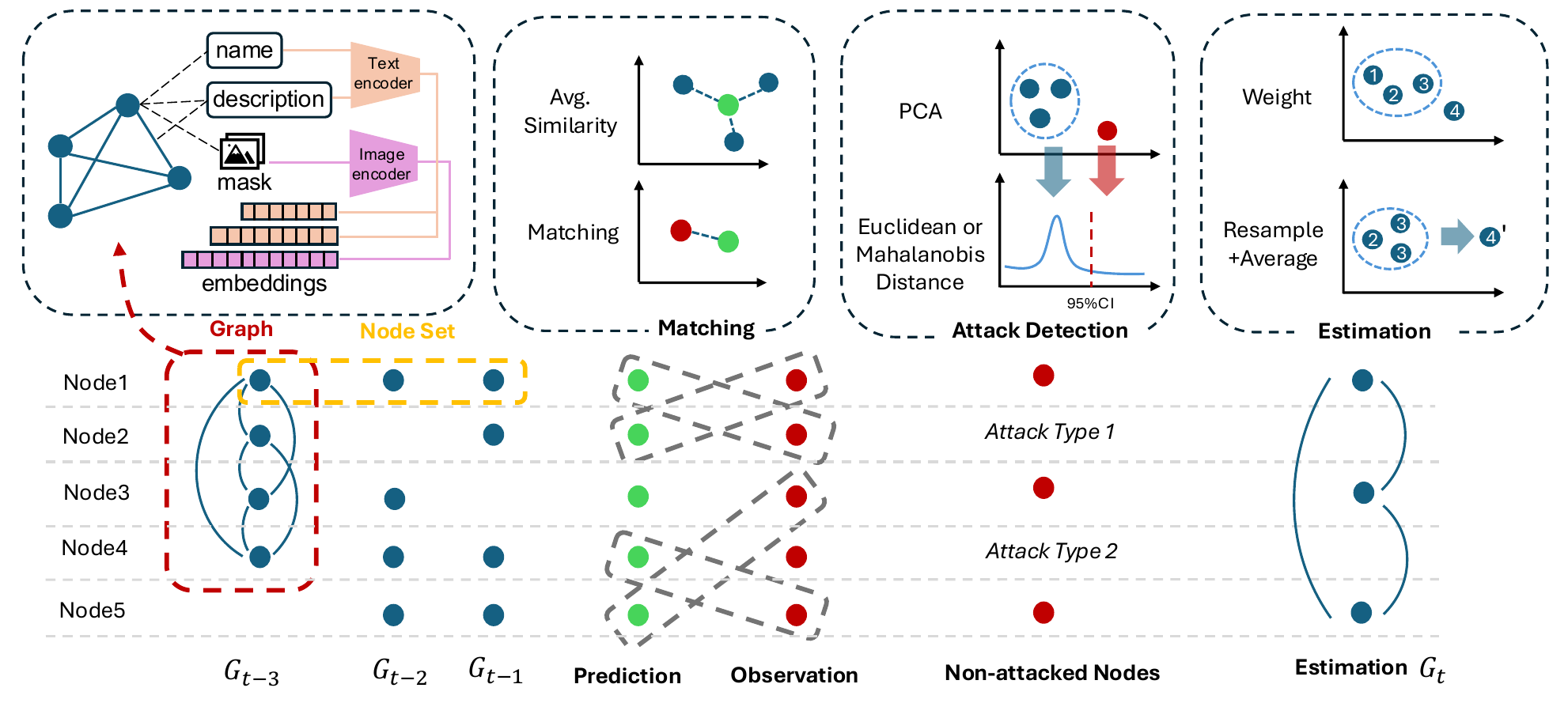}
    \caption{The particle filter-based attack detection framework consists of three main modules: matching, attack detection, and state estimation. }
    \label{fig: attack detection algorithm}
    \vspace{-5mm}
\end{figure*}

Our particle filter model tracks graph evolution and analyzes adversarial perturbations. It proceeds in three stages: prediction and matching, attack detection, and state estimation (Fig.~\ref{fig: attack detection algorithm}). 
Each entity (node or edge) is treated as a particle
At each timestep, particles match the incoming observations with prior particles. The attack detection algorithm applies statistical tests to newly matched particles and discards those marked as attacked. 
Finally, weighting, resampling, and estimation produce an updated, temporally consistent graph state, preserving persistent NodeSet/EdgeSet identities and the relational structure over time.

\vspace{-1mm}

\subsubsection{Prediction and Matching.}

To maintain persistent object identities across frames, we equip the NodeSet with the Prediction-Matching algorithm and use it to decide whether a newly detected Node belongs to an existing object list or creates a new one.

For any node \(v_i^t\) we denote  
\(\mathbf{n}(v)\) = name embedding,  
\(\mathbf{d}(v)\) = description embedding,  
\(\mathbf{f}(v)\) = visual feature embedding,  
all \(\in\mathbb{R}^{d_m}\).
Cosine similarity for modality $m\in\{n,d,f\}$ is defined as $\operatorname{sim}_m(u,v)$.



Let $\mathcal{A}_{t-1}$ be the set of NodeSet IDs that are still active at time $t-1$. Each set maintains an Active Level \(A_i^t \in [0,3]\) (initialized to \(3\)) indicating recent presence: if unmatched at time \(t\), \(A_i^t \leftarrow \max(0, A_i^{t-1}-1)\); if matched, \(A_i^t \leftarrow \min(3, A_i^{t-1}+1)\). When \(A_i^t=0\), the node is considered absent and ceases matching. For every active object ID \(i\in\mathcal{A}_{t-1}\) and each modality
\(m\in\{n,d,f\}\) compute the mean and standard deviation of pairwise similarities inside the NodeSet:

\vspace{-5mm}

\begin{equation}
\begin{aligned}
\mu_{i,m}^{\,t-1}
&= \frac{2}{|\mathcal{N}_{i,{t-1}}|(|\mathcal{N}_{i,{t-1}}|-1)}
   \sum_{\substack{p<q\\ v_i^p,v_i^q\in \mathcal{N}_{i,{t-1}}}}
   \operatorname{sim}_m(v_i^p,v_i^q),\\
\sigma_{i,m}^{\,t-1}
&= \mathrm{StdDev}\!\left\{
      \operatorname{sim}_m(v_i^p,v_i^q)\;\big|\;
      p<q,\; v_i^p,v_i^q\in \mathcal{N}_{i,t-1}
   \right\}.
\end{aligned}
\end{equation}

To match the observed node to NodeSet, given a fresh node \(v_j^t\) and a candidate NodeSet \(i\), define
\vspace{-3mm}
\begin{equation}
    s_{j,i,m}
   \;=\;
   \frac{1}{|\mathcal{N}_{i,t-1}|}
   \sum_{v\in\mathcal{N}_{i,t-1}}
      \operatorname{sim}_m(v_j^t,v),
   \quad
   m\in\{n,d,f\}.
\end{equation}

Candidate filter (\(\le 1\)std in every modality):
\begin{equation}
    \mathcal{C}_j
   \;=\;
   \Bigl\{
      i\in\mathcal{A}_{t-1}
      \;\Bigl|\;
      \bigl|s_{j,i,m}-\mu_{i,m}^{\,t-1}\bigr|
      \le
      \sigma_{i,m}^{\,t-1},
      \;
      \forall m\in\{n,d,f\}
   \Bigr\}.
\end{equation}

Finally, the assignment rule based on measuring the remaining candidate distance:
\begin{equation}
   \mu_t(v_{j,t})
   =
   \begin{cases}
      \displaystyle\arg\min_{i\in\mathcal{C}_j} \sum_{m\in\{n,d,f\}}
      \bigl|s_{j,i,m}-\mu_{i,m}^{\,t-1}\bigr|,
         & \mathcal{C}_j\neq\varnothing,\\[6pt]
      \text{new\_id}, & \mathcal{C}_j=\varnothing.
   \end{cases}
   \label{eq:sim-assign}
\end{equation}

If \(\mu_t(v_j^t)\) returns an existing ID, append \(v_j^t\) to that NodeSet and update \(A_i^{t}\);
otherwise, spawn a new NodeSet. Confirmed edges \(e_{ab}^t\) are then inserted into \(E_{\mu_t(a)\,\mu_t(b),t}\) as before. After matching, the NodeSet goes into the Attack detection process.

\subsubsection{Attack Detection.}

To detect text (Type 1) and visual (Type 2) modification attacks, those that manifest as anomalies in embedding space, we employ a dedicated statistical procedure. Because the same description embeddings are later reused when assessing EdgeSets, we illustrate the method with the NodeSet case; the extension to edges is entirely analogous. A NodeSet with a fresh match in frame \(t\) is denoted
\(\mathcal{N}_{i,t}=\mathcal{N}_{i,t-1}\cup\{v_i^t\}\).
Let  
\(
   \tilde{\mathcal{N}}_{i,t-1}
   \subseteq
   \mathcal{N}_{i,t-1}
\)
be the subset of nodes that \emph{have not} been flagged as attacks
in earlier frames; these constitute the “clean” reference set.

For each modality \(m\!\in\!\{d,f\}\)  
(description embedding, visual feature embedding), we use the Principal Component Analysis (PCA) to project it to a low-dimensional space, which removes noisy, low-variance directions. A PCA mapping on the clean reference data is computed by
\vspace{-1mm}
\begin{equation}
    \mathbf{W}_{i,m}^{\,t-1}
   \;=\;\operatorname{PCA}\!\bigl(
         \{\mathbf{m}(v)\}_{v\in\tilde{\mathcal{N}}_{i,t-1}},
         \;k_m
      \bigr),
\end{equation}

\noindent where \(k_m\) is chosen such that \(\ge\!99\%\) variance is retained.
the resulting embeddings are modeled as Gaussian distributions:
\vspace{-5mm}

\begin{equation}
   \mathbf{z}_{m}(v)
   = \mathbf{W}_{i,m}^{\,t-1\,\top}\mathbf{m}(v)
   \;\sim\;
   \mathcal{N}\!\bigl(
      \mathbf{c}_{i,m}^{\,t-1},
      \,\boldsymbol{\Sigma}_{i,m}^{\,t-1}
   \bigr)
\end{equation}



\noindent with the centroid and covariance of PCA-reduced embeddings
\begin{equation}
    \mathbf{c}_{i,m}^{\,t-1}
     = \frac{1}{|\tilde{\mathcal{N}}_{i,t-1}|}
       \sum_{v\in\tilde{\mathcal{N}}_{i,t-1}}
       \mathbf{z}_{m}(v), \hspace{1mm}\boldsymbol\Sigma_{i,m}^{\,t-1}
     = \mathrm{Cov}\!\bigl\{\mathbf{z}_{m}(v)\bigr\}.
 \end{equation}

Use the Mahalanobis distance for measuring Text and Visual Modification attack. For a new node \(v_i^t\) compute
\begin{equation}
    \mathcal{M}_{i,t}^{(m)}
   \;=\;
   \bigl(
      \mathbf{z}_{m}(v_i^t)-\mathbf{c}_{i,m}^{\,t-1}
   \bigr)^{\!\top}
   \bigl(\boldsymbol\Sigma_{i,m}^{\,t-1}\bigr)^{-1}
   \bigl(
      \mathbf{z}_{m}(v_i^t)-\mathbf{c}_{i,m}^{\,t-1}
   \bigr),
\end{equation}

Assume under the no-attack hypothesis, the squared Mahalanobis statistic $\mathcal{M}_{i,t}^{(\cdot)}\sim\chi^2_{k,\alpha}$, which is the $(1-\alpha)$-quantile of a chi-square distribution with \(k\) degrees of freedom. So flag the attack as
\begin{equation}
    v_i^t\;\text{as}
   \;\begin{cases}
       \text{Type 1 Attack(text)}  & \text{if } \mathcal{M}_{i,t}^{(d)}>\chi^2_{k_d,\alpha},\\[2pt]
       \text{Type 2 Attack(image)} & \text{if } \mathcal{M}_{i,t}^{(f)}>\chi^2_{k_f,\alpha}.
     \end{cases}
\end{equation}

For the structural attacks, specifically object obstruction (Type 3) and object injection (Type 4), the detector relies on lightweight temporal heuristics rather than statistical distance.  After each matching step, the algorithm keeps a two-frame sliding window of NodeSet activity.  If an object that previously carried a high task-importance score ($\pi\ge\pi_{\text{high}}$) fails to obtain a match in both the current and the immediately preceding frame, the system labels the disappearance an obstruction attack.  Conversely, any NodeSet that first appears within the last two frames and whose reasonability score falls below a predefined threshold ($\rho\le\rho_{\text{low}}$) is treated as an injection attack.

In the attack detection final step, if $v_i^t$ is flagged as attack, it is appended to the $\mathcal{N}_{i,t}$ but excluded from future Gaussian updates; otherwise going to the Node Estimation process, including weight and resample, then stored in a clean reference $\tilde{\mathcal{N}}_{i,t}$.

\subsubsection{Estimation.}\label{estimation module}

After a new observation \(v_{i,t}\) has been matched to object \(i\) and detected as a non-attack node, the particle filter estimates the new node by refining that observation by resampling historical detections that resemble it. The estimation proceeds as follows.

Let the newly matched observation of NodeSet $i$ at frame $t$ be
$v_{i,t}$ and denote by
\[
   \tilde{\mathcal{N}}_{i,t-1}
   =\{v^{(1)},\dots,v^{(M)}\}\subseteq\mathcal{N}_{i,t-1},
   \quad M\ge 5,
\]
the set of \emph{non-attack} detections accumulated up to
$t\!-\!1$.  
Considering each attribute as a particle, the filter updates $v_{i,t}$ by weighted resampling over this reference set.

For any scalar attribute \(a\in\{p,\pi,\rho\}\) (probability, importance, reasonability), let  
$a^\star = a(v_i^t)$ and
$a^{(j)} = a(v^{(j)})$ ($j=1\dots M$).  Define the weight 

\begin{equation}
    \sigma_a
      = \max\Bigl (\operatorname{Std}\{ a^{(1:M)}\},\;\sigma_{\min}\Bigr),
   \qquad
\end{equation}
\vspace{-3mm}
\begin{equation}
    w_j
      = \frac{\exp\!\bigl[-(a^{(j)}-a^\star)^2/(2\sigma_a^2)\bigr]}
             {\sum_{k=1}^{M}\exp\!\bigl[-(a^{(k)}-a^\star)^2/(2\sigma_a^2)\bigr]}.
\end{equation}
   
So $w_j$ is the normalized weight of the particle, based on its distance to the new particle. Similarly, for embedding attributes, modality \(m\in\{n,d,f\}\) (name, description, image) 
let
${m}^\star = {m}(v_i^t)$ and
${m}^{(j)} = {m}(v^{(j)})$.
Weight is defined as

\vspace{-5mm}

\begin{equation}
    sim_j
   = \frac{{m}^{(j)}\!\cdot\!{m}^\star}
          {\|{m}^{(j)}\|\,\|{m}^\star\|},
   \qquad
   w_j
   = \frac{\exp(\lambda_m sim_j)}
          {\sum_{k=1}^{M}\exp(\lambda_m sim_k)},
\end{equation}

\noindent with a scale factor $\lambda_m>0$.

Then use the weight to resample \(M\) indices
$s_1,\dots,s_M \sim \operatorname{Multinomial}(w_{1:M})$ yields estimated scalar particle and embedding particle:
\vspace{-5mm}

\begin{equation}
    \hat{a}
   \;=\;\frac{1}{M}\sum_{k=1}^{M} a^{(s_k)},\quad  \hat{{m}}
   \;=\;
   \frac{\sum_{k=1}^{M}{m}^{(s_k)}}%
        {\Bigl\|\sum_{k=1}^{M}{m}^{(s_k)}\Bigr\|},
\end{equation}

\noindent where $\hat{a}$ is the estimated new scalar particle.




Finally, replacing each raw attribute in $v_i^t$ by its resampled estimate gives the estimated node
\vspace{-0.2cm}
\begin{equation}
    \hat{v}_{i,t}
   =\bigl(
      \hat{p},\;\hat{\pi},\;\hat{\rho},\;
      \hat{{n}},\;\hat{{d}},\;\hat{{f}}
    \bigr),
\end{equation}
   
which becomes the particle filter’s estimation state for object $i$ at time $t$.
If $M<5$ (insufficient history), we skip resampling and simply set
$\hat{v}_{i,t}=v_{i,t}$.

%% file: sec/4_experiment.tex
\section{Experiment}

\begin{table*}[tb]
  \centering
  \caption{Attack-wise Accuracy and F1-score (\%) across scenes for frame-level and video-level baselines. Each attack type forms a block, with sub-rows indicating methods. Columns represent scene categories. CADAR outperforms VLM-based (GPT-5-mini, Gemini-2.5-flash) and supervised video models (3D-ResNet, ViViT) in 23 out of 25 attack–scene blocks.}
  \vspace{-0.2cm}
  \label{tab:attack_by_scene_var}
  \scriptsize 
  \setlength{\tabcolsep}{3pt}
  \renewcommand{\arraystretch}{0.6}


  \begin{tabular}{m{1.5cm}m{1cm}c|ccccc|c}

    \toprule
    \multicolumn{2}{c}{} &
      \multicolumn{6}{c}{\textbf{Class-specific Accuracy / F1-score (\%)}} \\
    \cmidrule(lr){4-9}
    \textbf{Attack} &  \textbf{Detection} &\textbf{Method} &
      \textbf{School\&Office} & \textbf{Traffic\&Mobility} & \textbf{Industrial\&Infra} &
      \textbf{Medical\&Lab} & \textbf{Public\&Home} &
      \textbf{Avg Acc/F1} \\
    \midrule

    \rowcolor{lightgray}
    \cellcolor{white}\multirow{5}{*}{\textbf{Text}} &
    \cellcolor{white}\multirow{3}{*}{\textbf{Frame}} &
    \textbf{CADAR (Ours)}      & \textbf{68.8~/~80.2} & \textbf{69.0~/~78.2} & \textbf{78.5~/~91.1} & \textbf{71.0~/~82.4} & \textbf{72.7~/~80.1} & \textbf{72.0~/~82.4} \\
     & & GPT-5-mini     & 50.6~/~67.2 & 59.6~/~74.7 & 52.1~/~68.5 & 52.5~/~68.9 & 61.1~/~75.9 & 55.2~/~71.0 \\
     & & Gemini-2.5-flash & 73.0~/~84.4 & 58.4~/~73.7 & 77.4~/~87.2 & 41.6~/~58.8 & 51.4~/~67.9 & 60.4~/~74.4 \\ 
     \cmidrule(lr){2-9}
      & \multirow{2}{*}{\textbf{Video}}
      & 3D ResNet  & 12.5~/~22.2  & 10.9~/~19.6 & 12.8~/~22.6 & 15.6~/~27.0 & 40.0~/~57.1 & 18.4~/~29.7 \\
     & & ViViT      & 26.3~/~41.7 & 70.6~/~82.8 & 25.0~/~42.1 & 33.3~/~50.0 & 43.5~/~60.6 & 39.7~/~55.4 \\
    \midrule

    \rowcolor{lightgray}
    \cellcolor{white}\multirow{5}{*}{\textbf{Visual}} &
    \cellcolor{white}\multirow{3}{*}{\textbf{Frame}} &
    \textbf{CADAR (Ours)}  & \textbf{80.3~/~92.1} & \textbf{79.2~/~85.2} & \textbf{67.2~/~78.9} & \textbf{69.7~/~81.0} & \textbf{70.4~/~84.5} & \textbf{73.4~/~84.3} \\
       & & GPT-5-mini     & 51.0~/~67.6 & 61.5~/~76.2 & 50.6~/~67.2 & 45.5~/~62.5 & 59.8~/~74.9 & 53.7~/~69.7 \\
      & & Gemini-2.5-flash & 49.5~/~66.3  & 76.7~/~86.8 & 43.9~/~61.0 & 58.7~/~74.0 & 40.8~/~58.0 & 53.9~/~69.2 \\
      \cmidrule(lr){2-9}
      & \multirow{2}{*}{\textbf{Video}}
      & 3D ResNet  & 16.5~/~28.3  & 7.2~/~13.3 & 8.3~/~15.3 & 12.7~/~22.5 & 8.3~/~15.4 & 10.6~/~19.0 \\
     & & ViViT      & 22.2~/~36.4 & 44.4~/~61.5 & 20.0~/~33.3 & 28.6~/~44.4 & 25.0~/~40.0 & 28.0~/~43.1 \\
    \midrule

    \rowcolor{lightgray}
    \cellcolor{white}\multirow{5}{*}{\textbf{Obstruction}} &
    \cellcolor{white}\multirow{3}{*}{\textbf{Frame}} &
    \textbf{CADAR (Ours)}      & \textbf{78.2~/~87.3} & \textbf{85.1~/~91.9} & \textbf{81.2~/~89.3} & \textbf{78.3~/~89.1} & \textbf{80.7~/~92.1} & \textbf{80.7~/~89.9} \\
      & & GPT-5-mini     & 75.4~/~85.9 & 46.0~/~63.0 & 61.7~/~76.3 & 46.3~/~63.3 & 63.2~/~77.4 & 58.5~/~73.2 \\
      & & Gemini-2.5-flash & 64.7~/~78.6 & 51.6~/~68.1 & 55.0~/~71.0 & 36.9~/~53.9 & 43.6~/~60.7 & 50.4~/~66.5 \\
      \cmidrule(lr){2-9}
      & \multirow{2}{*}{\textbf{Video}}
      & 3D ResNet  & 8.1~/~15.0  & 3.0~/~5.9 & 5.0~/~9.5 & 9.4~/~17.2 & 4.5~/~8.7 & 6.0~/~11.3  \\
     & & ViViT      & 50.0~/~66.7 & 54.5~/~70.6 & 46.2~/~63.2 & 71.4~/~83.3 & 41.7~/~42.3 & 52.8~/~65.2 \\
    \midrule

    \rowcolor{lightgray}
    \cellcolor{white}\multirow{5}{*}{\textbf{Injection}} &
     \cellcolor{white}\multirow{3}{*}{\textbf{Frame}} &
      \textbf{CADAR (Ours)}      & \textbf{76.9~/~81.1} & \textbf{81.4~/~89.5} & \textbf{72.2~/~78.9} & \textbf{73.5~/~79.0} & \textbf{74.0~/~82.6} & \textbf{75.6~/~82.2} \\
      & & GPT-5-mini     & 69.1~/~81.7 & 60.7~/~75.6 & 56.3~/~72.0 & 55.2~/~71.2 & 63.3~/~77.5 & 60.9~/~75.6 \\
      & & Gemini-2.5-flash & 70.1~/~82.4 & 54.1~/~70.2 & 62.8~/~77.2 & 61.5~/~76.1 & 53.1~/~69.4 & 60.3~/~75.0 \\
      \cmidrule(lr){2-9}
      & \multirow{2}{*}{\textbf{Video}}
      & 3D ResNet  & 6.6~/~12.3& 12.8~/~22.7 & 9.8~/~17.8 & 13.3~/~23.4 & 15.7~/~27.2 & 11.6~/~20.7 \\
     & & ViViT      & 30.0~/~46.2 & 45.5~/~62.5 & 20.0~/~33.3 & 25.0~/~40.0 & 55.6~/~71.4 & 35.2~/~50.7 \\
    \midrule

    \rowcolor{lightgray}
    \cellcolor{white}\multirow{5}{*}{\textbf{Non attack}} &
     \cellcolor{white}\multirow{3}{*}{\textbf{Frame}} &
      \textbf{CADAR (Ours)}      & \textbf{75.3~/~80.8} & \textbf{68.9~/~80.2} & \textbf{71.6~/~84.9} & \textbf{74.6~/~85.2} & \textbf{67.2~/~79.4} & \textbf{71.5~/~82.1} \\
     & & GPT-5-mini & 71.8~/~83.6 & 52.1~/~68.5& 62.8~/~77.1 & 70.5~/~82.7 & 68.1~/~81.0 & 65.1~/~77.4 \\
      & & Gemini-2.5-flash & 54.2~/~70.3 & 38.2~/~55.3 & 44.0~/~61.1 & 41.1~/~58.3 & 44.1~/~61.2 & 44.3~/~61.2 \\
      \cmidrule(lr){2-9}
      & \multirow{2}{*}{\textbf{Video}}
      & 3D ResNet  & 8.5~/~15.7 & 11.7~/~20.9 & 8.6~/~15.9 & 15.9~/~27.5 & 29.4~/~45.5 & 14.8~/~25.1 \\
      & & ViViT      & 40.0~/~57.1  & 15.4~/~26.7 & 7.7~/~14.3 & 40.0~/~57.1 & 30.8~/~47.1 & 26.8~/~40.5 \\
    \midrule

    \multicolumn{2}{l}{\textbf{Overall Mean (per method):}} &
      \multicolumn{6}{c}{%
        \textbf{CADAR: 74.6 / 84.2} \quad
        GPT-5-mini: 58.7 / 73.4 \quad
        Gemini-2.5-flash: 53.9 / 69.3 \quad
        3D ResNet: 12.3 / 21.1 \quad
        ViViT: 36.5 / 51.0
      }\\
    \bottomrule
  \end{tabular}
  \vspace{-2mm}
\end{table*}

\begin{table}[ht]
\centering
\caption{Recall comparison for hybrid-attack scenarios(\%), CADAR achieves the highest recall across both hybrid combinations.}
\setlength{\tabcolsep}{2pt} 
\scriptsize 
\renewcommand{\arraystretch}{0.6} 
  
\begin{tabular}{lcccc}
\toprule
\multirow{2}{*}{\textbf{Hybrid Attack Type}} & \multicolumn{4}{c}{\textbf{Recall of Method (\%)}} \\ 
\cmidrule(lr){2-5}
 & GPT-5-mini & Grok-4-fast & \textbf{CADAR} & Gemini-2.5-flash \\
\midrule
Text+Visual               & 35.0 & 41.2 & \textbf{76.4} & 29.9 \\
Obstruction+Injection     & 32.7 & 26.1 & \textbf{80.4} & 45.8 \\
\bottomrule
\end{tabular}
\label{tab:hybrid_attack}
\end{table}

\begin{table}[ht]
\centering
\caption{Recall (\%) of attack detection under different numbers of simultaneous attacks. CADAR maintains strong performance as the number of attacks increases, whereas baseline models degrade substantially.}
\setlength{\tabcolsep}{2pt}
\scriptsize 
\renewcommand{\arraystretch}{0.6}
\begin{tabular}{lccccc}
\toprule
\multirow{2}{*}{\textbf{Method}}
& \multicolumn{4}{c}{\textbf{Number of Attacks}} 
& \multirow{2}{*}{\textbf{Avg}} \\
\cmidrule(lr){2-5}
& \textbf{1 Attack} & \textbf{2 Attacks} & \textbf{3 Attacks} & \textbf{4 Attacks} & \\
\midrule
GPT5-mini        & 66.4 & 40.2 & 31.5 & 29.6 & 41.9\\
Grok-4-fast      & 69.7 & 55.0 & 51.3 & 36.5 & 53.1\\
Gemini-2.5-flash & 61.7 & 43.6 & 33.0 & 31.5 & 42.5\\
\rowcolor{lightgray}
\textbf{CADAR}   & \textbf{81.0} & \textbf{78.3} & \textbf{75.2} & \textbf{67.2} & \textbf{75.4}\\
\bottomrule
\end{tabular}
\label{tab:multi_attack}
\vspace{-3mm}
\end{table}

In this section, we provide a detailed description of the \emph{CADAR} experimental settings used for performance evaluation. We then present and analyze the numerical results of our experiments, demonstrating that \emph{CADAR} significantly outperforms several baselines. An ablation study is also conducted to analyze the contributions of our key components.

\subsection{Experimental Settings}

\paragraph{CADAR-50K Dataset.}

\begin{figure}[t]
    \centering
    \includegraphics[width=3.4in]{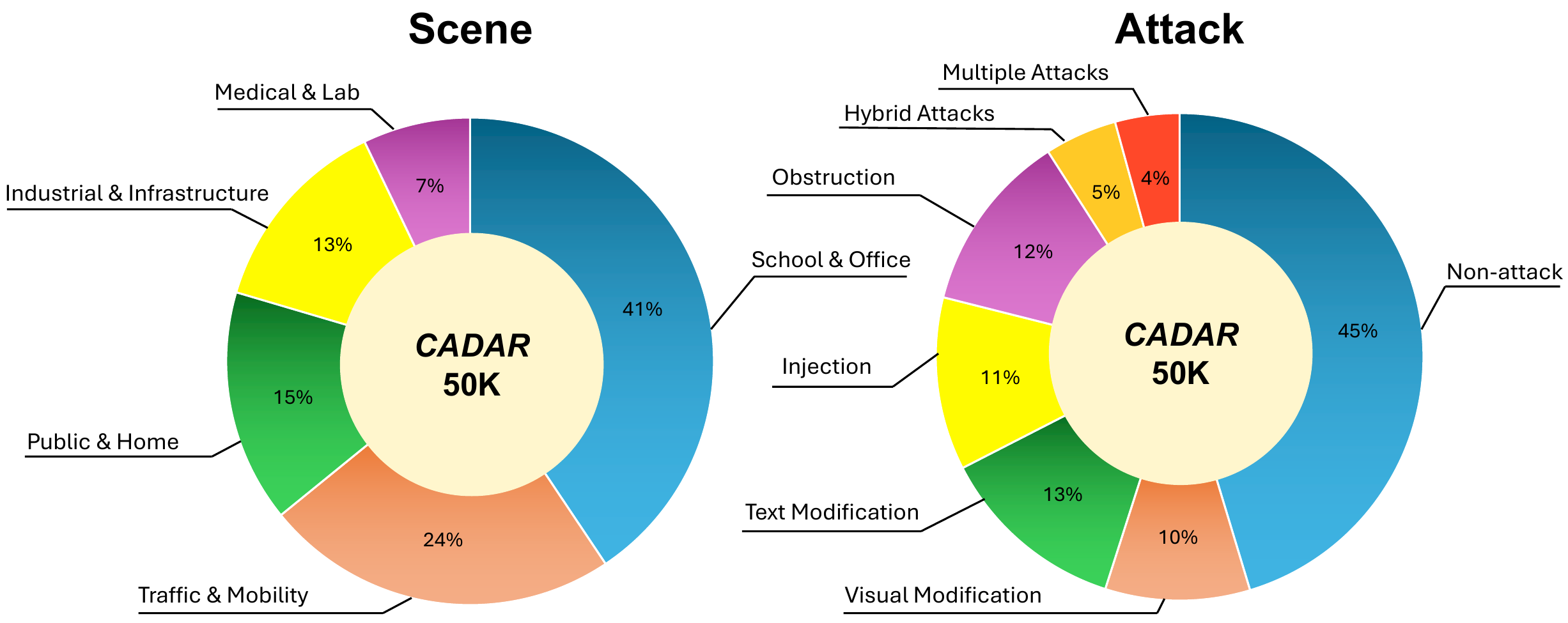}
    \vspace{-0.2in}
    \caption{CADAR-50K dataset overview.}
        \vspace{-0.2in}
    \label{fig:dataset}
\end{figure}

\begin{figure*}[t]
    \centering
    \includegraphics[width=7in]{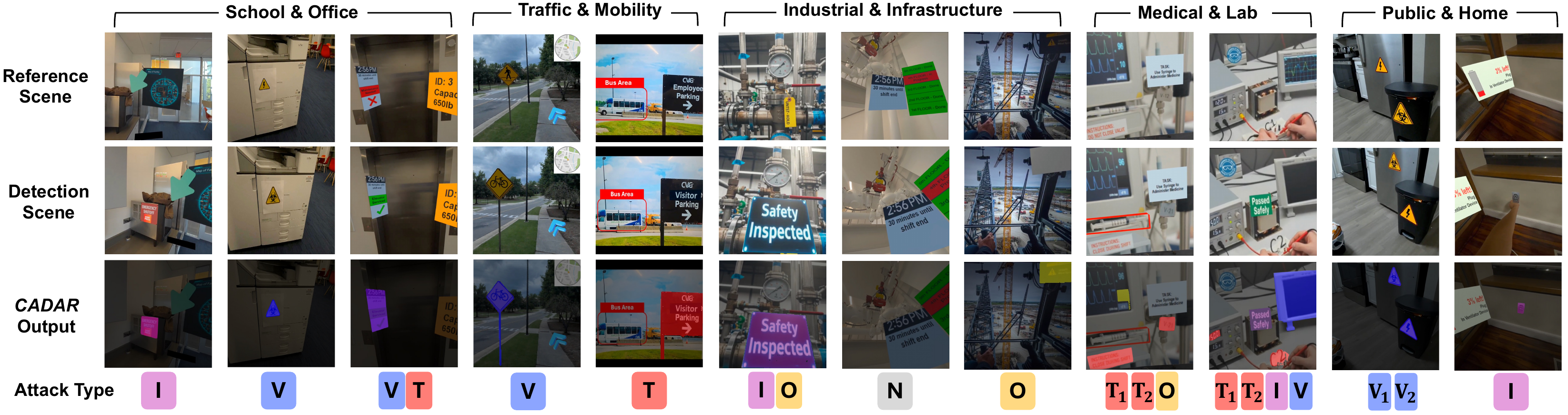}
    \caption{Visualization frame examples of \emph{CADAR} outputs across five scene categories in CADAR-50K 375 videos. Rows show the Reference Scene, Detection Scene, and \emph{CADAR} Output. Output attack types: I = Injection, V = Visual, T = Text, O = Obstruction, N = Non-attack. Some examples contain hybrid or multiple attacks.}
    \label{fig:visualization}
    \vspace{-3mm}
\end{figure*}

The CADAR-50K dataset extends the AR-VIM dataset~\cite{11202433}, which was designed to evaluate visual information manipulation in AR environments. CADAR-50K is the first publicly available dataset for AR cognitive-attack analysis, containing 375 videos across diverse scenes and more than 50K frames, covering all seven categories (Fig.~\ref{fig:dataset} and Fig.~\ref{fig:visualization}). Each sample is an 8–12 second AR-captured video (recorded using a Meta Quest 3 or Samsung Galaxy S24), where virtual content is introduced at a specific frame to induce an attack. For the non-attack elements, we use irrelevant background objects, benign AR elements, and semantically benign prompts that can interfere with VLM outputs but do not alter the human-perceived task semantics.

Considering scene availability, the dataset was collected using two complementary methods: (1) a virtual-scene method, where fully synthetic environments are first generated and displayed on a monitor or VR headset, and AR content is then overlaid onto these virtual scenes before being captured through video recording. This enables rapid, controllable creation of diverse attack scenarios; and(2) a real-world method, where AR headsets record first-person views in physical environments, introducing natural noise, lighting variation, and depth cues.
\vspace{-3mm}

\paragraph{Task, Baselines and Metrics.}

We evaluate our model on the task of video-based cognitive attack classification. The \emph{CADAR} dataset supports frame-wise detection, variable-length multi-label outputs, and localization of simultaneous attacks. We compare \emph{CADAR} against two categories of baselines that similarly accept multiple reference frames as input and output attack classes: (1) multimodal VLMs, including state-of-the-art \emph{OpenAI GPT-5}, \emph{Google Gemini-2.5}~\cite{comanici2025gemini} and \emph{xAI Grok-4}, which perform fine-grained frame-level detection; and (2) supervised video models, the CNN-based \emph{3D-ResNet}~\cite{tran2018closer} and transformer-based \emph{ViViT}~\cite{arnab2021vivit}, which output a single prediction per video. 

In our first experiment(Table~\ref{tab:attack_by_scene_var}), we use class-specific accuracy and F1-score to assess overall detection performance. 
 The class-specific accuracy is defined as $\mathrm{TP}/(\mathrm{TP}+\mathrm{FN}+\mathrm{FP})$, which penalizes both missed detections and false alarms for that class while ignoring true negatives that dominate under severe class imbalance.
For hybrid attack or multiple attack videos(Table~\ref{tab:hybrid_attack} and \ref{tab:multi_attack}), a prediction is considered correct only if all attack types are simultaneously identified. We also report per-class recall , $\mathrm{TP}/(\mathrm{TP}+\mathrm{FN})$, to measure correctness conditioned on the presence of the class.

\subsection{Numerical Results and discussion}


\paragraph{Cross-Scene Cognitive Attack Classification.}The results in Table~\ref{tab:attack_by_scene_var} demonstrate that \emph{CADAR} achieves the highest overall accuracy(74.6\%) and F1-score(84.2\%) and its average accuracy exceeding the next best baseline models (GPT-5-mini) by 16\%. CADAR attains the highest class-specific accuracy and F1 in 23 out of 25 attack–scene blocks. The margin is especially large for Visual and Obstruction attacks, where \emph{CADAR} delivers strong gains, underscoring its robustness to fine-grained semantic distortions.
We also observe that VLM baselines exhibit substantial accuracy fluctuations across different scenes, indicating that their reasoning processes are unstable and prone to hallucination. Because they rely on jointly processing all textual and visual inputs, they become increasingly sensitive to misleading cues, malicious prompts, and complex visual environments. This vulnerability highlights the advantage of our neuro-symbolic approach, which offers more controlled and explainable reasoning under diverse AR scenarios.

On the other hand, traditional video-classification models (3D-ResNet and ViViT) perform poorly across all attack types, suggesting that the models are inadequate for capturing the fine-grained semantic inconsistencies central to cognitive attack detection. Their limited performance is further exacerbated by the lack of large-scale, AR-specific training data, making these models particularly weak in zero- or few-shot scenarios.
 
Overall, the results indicate that \emph{CADAR}’s neurosymbolic design and temporal reasoning components provide strong advantages for detecting subtle, localized, and multi-instance AR cognitive attacks, achieving robust performance across diverse environments and attack conditions.

\vspace{-2mm}

\paragraph{Hybrid and Multiple Attacks}

 Table \ref{tab:hybrid_attack} and \ref{tab:multi_attack} evaluate model robustness in hybrid and multi-attack settings. In hybrid-attack scenarios, \emph{CADAR} attains the highest recall for both Text+Visual(76.4\%) and Obstruction+Injection(80.4\%) cases and surpasses all VLM baselines by a substantial margin. This demonstrates that the proposed neurosymbolic graph reasoning framework can effectively detect the attacks separately.
In multiple attack scenarios, the performance of VLM baselines declines sharply as attack complexity grows, whereas \emph{CADAR} retains strong recall, from 81.0\% for a single attack to 67.2\% for four simultaneous attacks, showing the smallest performance drop among all compared methods. These results highlight \emph{CADAR}’s resilience in multi-attack scenarios and its clear advantage in detecting objects in parallel.

\paragraph{Effect of Scene Complexity.}

The recall trends in Fig.~\ref{fig: ablation study}(a) reveal a clear divergence between VLM-based baselines and \emph{CADAR} as scene complexity increases. For GPT-5-mini and Gemini-2.5-flash, recall drops sharply when the number of key objects and frames increases, suggesting that conventional VLMs struggle with reasoning over long visual and textual context and are easily affected by irrelevant or distracting content. In contrast, \emph{CADAR} performance remains stable as reference frames increase, since it extracts complex scene information into a structural spatial-temporal perception graph. \emph{CADAR} improves its decision accuracy by accumulating prior evidence through additional reference frames, and evaluates objects and attacks separately rather than relying on a single holistic reasoning step. The \emph{CADAR} design results in substantially smaller performance degradation compared with standard VLMs.

\subsection{Ablation Study}

\paragraph{Estimation Module.}

\begin{figure}[t]
    \centering
    \includegraphics[width=3.3in]{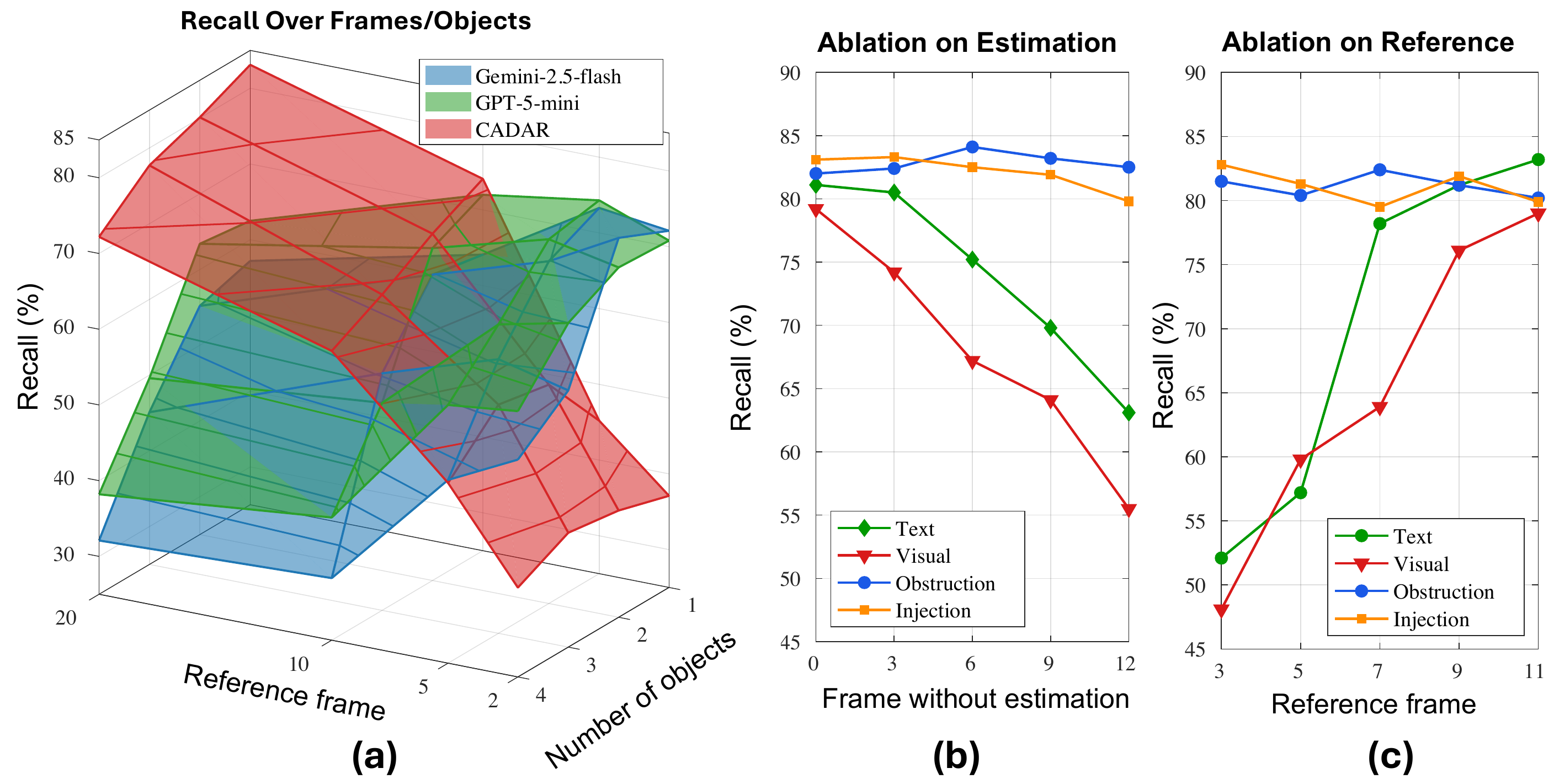}
    \vspace{-0.3in}
    \caption{Recall under increasing reference frames and object quantities (a), and ablation studies on the estimation module (b) and the number of reference frames (c).}
     \vspace{-3mm}
    \label{fig: ablation study}
\end{figure}

Fig.~\ref{fig: ablation study}(b) shows our ablation study examining the impact of the estimation module in particle filtering. When a polluted node that might have bad data due to VLM unstable output or due to an undetected attack, the estimation module's weighting and resampling can effectively prevent it from corrupting the ground truth knowledge. We randomly sample 6 videos per base attack type (24 in total) from the test split for the ablation study. As shown in the left plot of the figure, for text and visual modification attacks, the recall drops significantly as the number of frames without the estimation module increases. Specifically, the recall for text attacks decreases from approximately $81\%$ to $63\%$ and from approximately $79\%$ to $55\%$ for visual attacks. This indicates that the estimation module is crucial for correctly identifying and mitigating errors in more subtle attack types. In contrast, for obstruction and injection attacks, the recall remains relatively stable, suggesting these attacks are more straightforward to detect and less dependent on the estimation module's corrective capabilities.
\vspace{-2mm}
\paragraph{Reference Frames.}
Fig.~\ref{fig: ablation study}(c) shows the impact of the number of reference frames on recall. A larger number of reference frames, or more prior knowledge of the scene, will result in higher recall. For text and visual modification attacks, increasing the number of reference frames from 3 to 11 significantly improves performance. The recall for text attacks increases from around $52\%$ to $83\%$ and from around $48\%$ to $79\%$ for visual attacks. This demonstrates that a larger set of reference frames provides more prior knowledge for more accurate statistical reasoning, which is essential for detecting subtle text and visual changes. Similar to the estimation module study, the recall for obstruction and injection attacks remains largely unaffected by the number of reference frames, consistently staying around $80\%$. This reinforces the finding that these attacks are less sensitive to the amount of contextual information and prior knowledge.





%% file: sec/5_conclusion.tex
\section{Conclusion}
We presented \emph{CADAR}, a neuro-symbolic framework that couples a perception graph with particle-filter reasoning to yield interpretable attack detection in AR/MR. On the CADAR-50K dataset, \emph{CADAR} performs strongly across all attack types, and ablations highlight the value of our estimation module and richer reference histories. Overall, \emph{CADAR} highlights the value of neuro-symbolic design by combining the adaptability of foundation models with symbolic reasoning for a robust and interpretable AR safety mechanism.

%% file: sec/6_appendix.tex
\appendix
\section*{Appendix}

\section{CADAR Implementation Details}

This section provides the full implementation details of \emph{CADAR}, covering perception-graph construction, multimodal embedding generation, temporal reasoning, and statistical attack detection. The system follows a modular architecture implemented across the \texttt{Node.py}, \texttt{Edge.py}, \texttt{Graph.py}, \texttt{NodeSet.py}, \texttt{EdgeSet.py}, and \texttt{Utils.py} modules in our codebase.

\subsection{Perception Graph Construction}

\paragraph{Object Detection and Segmentation.}
For each frame, \emph{CADAR} extracts object proposals using two vision modules:
\begin{itemize}
    \item Zero-shot object detection: 
    
    \texttt{google/owlv2-base-patch16-ensemble}~\cite{minderer2024scalingopenvocabularyobjectdetection}, which produces category-agnostic bounding boxes and confidence scores.
    \item Segmentation masks: \texttt{SAM2}~\cite{ravi2024sam2segmentimages}, used to obtain object-level silhouettes for robust visual embedding extraction.
\end{itemize}

\paragraph{Textual Attribute Generation.}
To obtain semantic descriptions associated with the detected region, we use a lightweight VLM (GPT-4.1) generates short, structured textual descriptions. These descriptions include attributes such as object type, color, role, contextual relevance, and scene relationships. The attributes also include numerical importance, reasonability and probability.

\paragraph{Multimodal Embedding Extraction.}
Each detected object is represented by three embedding modalities, each extracted using a specialized encoder:
\begin{itemize}
    \item Name embedding: \texttt{paraphrase-MiniLM-L6-v2} (384-d).
    \item Description embedding: \texttt{all-mpnet-base-v2} (768-d).
    \item Visual embedding: a \texttt{ResNet-34} feature encoder applied to SAM2-extracted object crops (512-d).
\end{itemize}

These multimodal perceptual vectors are then projected using PCA for statistical reasoning (described below).

\paragraph{Node and Edge Formation.}
\paragraph{Implementation Details.}
The software architecture is structured around two primary classes, \texttt{Node.py} and \texttt{Edge.py}, which manage the lifecycle and serialization of the perception graph components. The \texttt{Node} class orchestrates the bottom-up extraction pipeline, initializing the vision models to generate object masks and invoking the VLM to populate attribute fields (e.g., reasonability, importance) while computing and storing dense multimodal embeddings. Complementing this, the \texttt{Edge} class executes the relational analysis by ingesting the instantiated node list and source image; it prompts the LLM to infer pairwise dependencies (such as spatial or functional links) and encodes these descriptions into vector format, thereby establishing the graph's connectivity before serializing the complete state to compressed binary files.

\subsection{Graph Backbone and Temporal Tracking}

\paragraph{Graph Management.}
The \texttt{Graph.py} module serves as the fundamental container for encapsulating the state of the perception system at discrete timestamps. It aggregates detected entities and their pairwise relationships into a coherent network structure, storing lists of \texttt{Node} and \texttt{Edge} instances alongside temporal metadata and references to the source imagery. Beyond acting as a static data structure, this module facilitates the serialization of the complete graph state into compressed binary formats to ensure efficient persistence and provides visualization utilities that render the probabilistic attributes and topological connectivity of the scene using network layout algorithms.

\paragraph{NodeSet and EdgeSet Evolution.}
To transform isolated frame-level detections into continuous historical tracks, the system relies on the \texttt{NodeSet.py} and \texttt{EdgeSet.py} modules. The \texttt{NodeSet} class maintains the identity of objects over time by associating new detections with existing tracks via multi-modal embedding similarity (visual, semantic, and nominal) rather than simple spatial overlap. It implements a particle filter-based mechanism to estimate stable node attributes from noisy observations and employs statistical methods—such as PCA and Mahalanobis distance—to detect anomalies or adversarial attacks. Complementing this, the \texttt{EdgeSet} tracks the evolution of semantic and spatial relationships between these persistent entities, preserving the history of interactions to support high-level reasoning as the scene evolves.

\subsection{Statistical Anomaly Detection}

\paragraph{Dimensionality Reduction and Feature Whitening.}
To ensure robust distance metrics in high-dimensional spaces, we apply Principal Component Analysis (PCA) to the embedding history of each track. For a set of historical embedding vectors $\mathbf{X} \in \mathbb{R}^{N \times D}$, we compute a projection that preserves a specific variance ratio. Specifically, for node textual and visual embeddings, we retain 99\% of the variance ($v=0.99$), while for edge description embeddings, we retain 90\% ($v=0.90$). For visual embeddings, we additionally apply whitening to normalize the principal components to unit variance. This projection results in a dynamic latent dimension $k$ (typically $16 \le k \le 32$), which defines the degrees of freedom for subsequent statistical tests.

\paragraph{Regularized Mahalanobis Distance.}
We model the distribution of "clean" reference frames for any given object or relationship as a multivariate Gaussian. To assess whether a new observation $x_t$ belongs to this distribution, we calculate the squared Mahalanobis distance. To prevent numerical instability when the covariance matrix is singular (common in short history windows), we introduce a regularization term $\epsilon \mathbf{I}$:
\begin{equation}
   d^2(x_t) = (x_t - \mu)^\top (\Sigma + \epsilon \mathbf{I})^{-1} (x_t - \mu) 
\end{equation}

where $\mu$ is the mean of the historical embeddings, $\Sigma$ is the covariance matrix, and $\epsilon = 10^{-6}$ is a smoothing factor.

\paragraph{Chi-Square Hypothesis Testing.}
Anomalies are identified by comparing the squared distance against a critical value derived from the Chi-square distribution with $k$ degrees of freedom. An observation is flagged as an attack (e.g., semantic modification or visual spoofing) if:
\begin{equation}
    d^2(x_t) > \chi^2_{1-\alpha,\,k}
\end{equation}

where $\alpha$ represents the significance level. We employ adaptive sensitivity thresholds: $\alpha=0.10$ for textual deviations, $\alpha=0.05$ for visual deviations, and $\alpha=0.01$ for edge relationship deviations. This statistical framework allows \emph{CADAR} to detect subtle adversarial manipulations that deviate from the established semantic baseline.

\subsection{Particle-Filter-Based Temporal Reasoning}

To mitigate the stochasticity of VLM outputs and detection noise, we implement a particle filter that maintains a belief state over node attributes. Unlike standard tracking which estimates position, our filter estimates semantic consistency.

\paragraph{Weight Update.}
For a node $n$ with history $H$, we define the weight $w_i$ of the $i$-th particle (attribute hypothesis) based on its similarity to the incoming observation $z_t$.
For scalar attributes (probability, importance, reasonability), weights follow a Gaussian kernel:
\begin{equation}
    w_i \propto \exp\left( -\frac{(v_i - z_t)^2}{2\sigma^2} \right)
\end{equation}

For high-dimensional embedding attributes, weights are derived from cosine similarity, scaled by a confidence factor $\lambda$:
\begin{equation}
    w_i \propto \exp\left( \lambda \cdot \frac{\mathbf{e}_i \cdot \mathbf{z}_t}{\|\mathbf{e}_i\| \|\mathbf{z}_t\|} \right)
\end{equation}

where $\lambda=1$ is the scaling factor used to tune the confidence in the top matches.

\paragraph{Resampling and Estimation.}
We perform importance resampling to filter outlier descriptions generated by the VLM. The final estimated state $\hat{x}_t$ is computed as the weighted centroid of the resampled population, effectively smoothing the semantic trajectory of the object and rejecting transient classification errors.

\begin{table*}[h]
\centering
\caption{Hyper-parameters for Statistical Detection and Tracking.}
\label{tab:hyperparams}
\begin{tabular}{lcc}
\toprule
\textbf{Parameter} & \textbf{Value} & \textbf{Description} \\
\midrule
Matching Threshold & 0.2 & Similarity floor for node association \\
Min. History Window & 5 frames & Minimum samples required for PCA/Mahalanobis \\
PCA Variance (Nodes) & 0.99 & Variance retained for text/image embeddings \\
PCA Variance (Edges) & 0.90 & Variance retained for edge descriptions \\
Regularization ($\epsilon$) & $10^{-6}$ & Covariance matrix stabilization term \\
$\alpha$ (Text) & 0.10 & Significance level for node text anomalies \\
$\alpha$ (Image) & 0.05 & Significance level for visual appearance anomalies \\
$\alpha$ (Edge) & 0.01 & Significance level for relationship anomalies \\
Particle Scale ($\lambda$) & 1.0 & Exponential scaling for embedding similarity weights \\
\bottomrule
\end{tabular}
\end{table*}

\subsection{Supporting Utilities}

\paragraph{Utility and Pre-processing.}
The \texttt{utils.py} module serves as a comprehensive toolkit for low-level visual processing and mathematical operations underpinning the higher-level graph logic. 

\subsection{Summary}

Overall, \emph{CADAR} integrates zero-shot perception, multimodal symbolic representation, PCA-based statistical modeling, and particle-filter temporal reasoning into a unified framework for robust AR cognitive attack detection. The modular architecture allows flexible substitution of detectors, encoders, and reasoning mechanisms, while ensuring reproducibility across all experimental settings.

\section{CADAR-50K Dataset Samples}
Representative sample videos for each class (excluding the No Attack category) is shown below. The full dataset consists of 375 videos across diverse scenes.

\begin{figure*}[ht]
    \centering
    \includegraphics[width=5.5in]{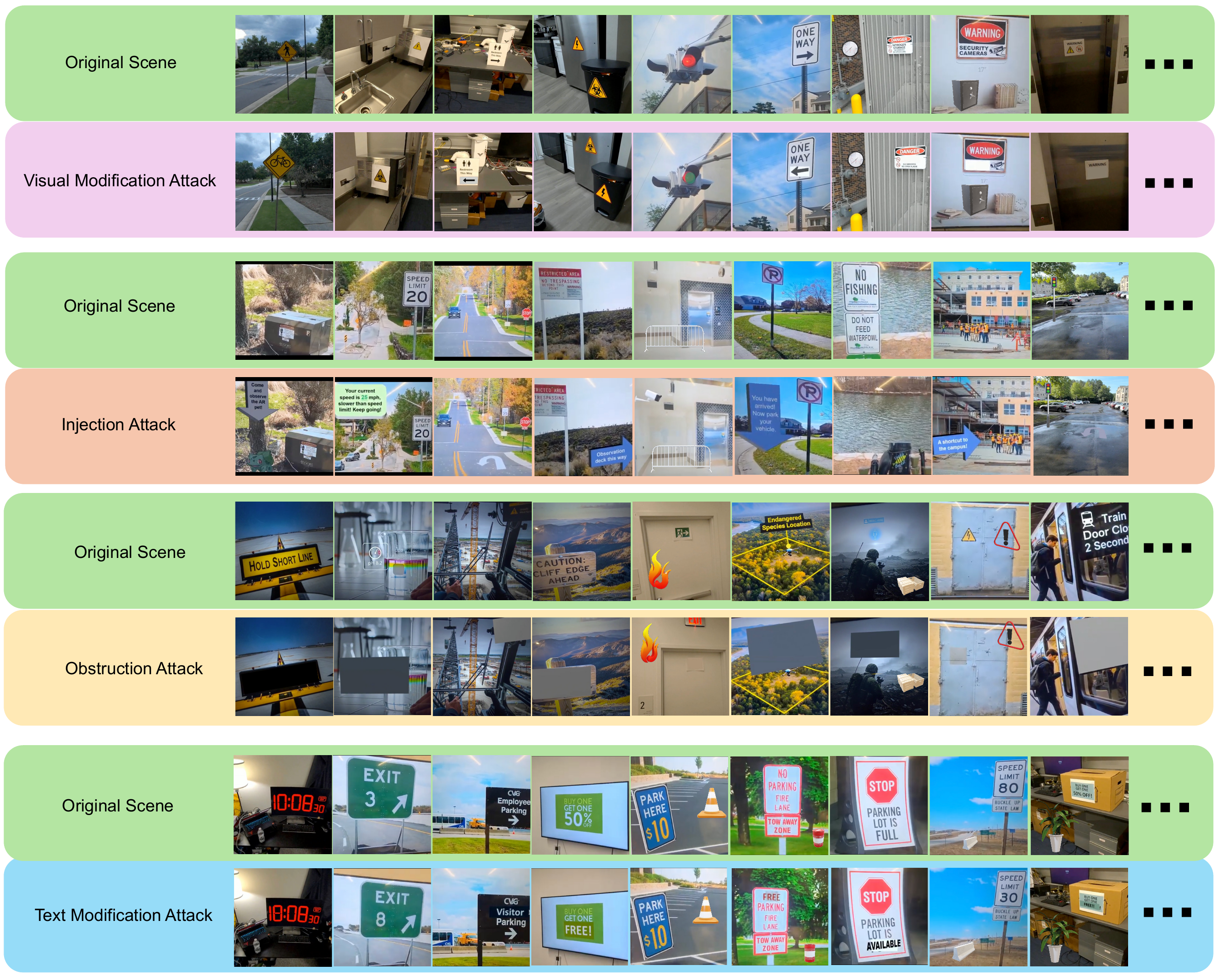} 
    \caption{Sample dataset content for each AR Attack Class. For video examples, please refer to \textbf{CADAR\_demo.mp4} in the supplementary materials.}
\end{figure*}

\subsection{Dataset Construction}

For the real-world scenes in the CADAR-50K dataset, we collected additional samples using a Meta Quest 3 headset and a cellphone camera. The video dataset contains both malicious cognitive attacks and benign augmented reality content. All data was recorded using the built-in pass-through camera of a Meta Quest headset and a Samsung Galaxy S24. We employed two distinct data generation pipelines to ensure diversity in environmental contexts and lighting conditions.

\subsection{In-Situ Augmented Reality Overlay}
The first method involved augmenting real-world environments with manipulated overlays to simulate attacks in situ. We first identified target objects within a physical scene (e.g., a store sign displaying a discount). A high-resolution photograph of the target was captured and subjected to digital manipulation. For instance, altering text from "50\% off" to "150\% off" to simulate a text manipulation attack.

The manipulated asset was imported into the Unity engine and mapped onto a planar mesh (Quad). We utilized Unity's scaling and transformation tools to align the virtual overlay with the physical object's dimensions and perspective in the real world. To introduce benign confounders, we analyze the scene context and suggest appropriate non-malicious objects. Based on these suggestions, synthetic 3D or 2D assets were generated and anchored within the scene. During the recording phase, a controller input triggered the visibility of the overlay, allowing us to capture the transition between the benign state and the attack state dynamically.

\subsection{Synthetic Scene Projection}
To expand the dataset beyond accessible physical locations, we utilized a fully synthetic generation pipeline. In the Unity environment, we instantiated a large-scale Quad (approximately $5m \times 5m$) to serve as a dynamic backdrop. 

We utilized a generative image model to synthesize realistic background scenes. The generation of scenes containing specific target objects is required for occlusion or manipulation attacks. Once the generated scene was applied to the backdrop, attack artifacts were positioned in the foreground. Although the background was two-dimensional, we achieved a pseudo-three-dimensional effect by strategically angling the foreground objects and the camera perspective. Similar to the in-situ method, the attack artifacts were toggled via controller input during recording to capture paired data samples.

\begin{figure}[ht]
    \centering
    \includegraphics[width=3in]{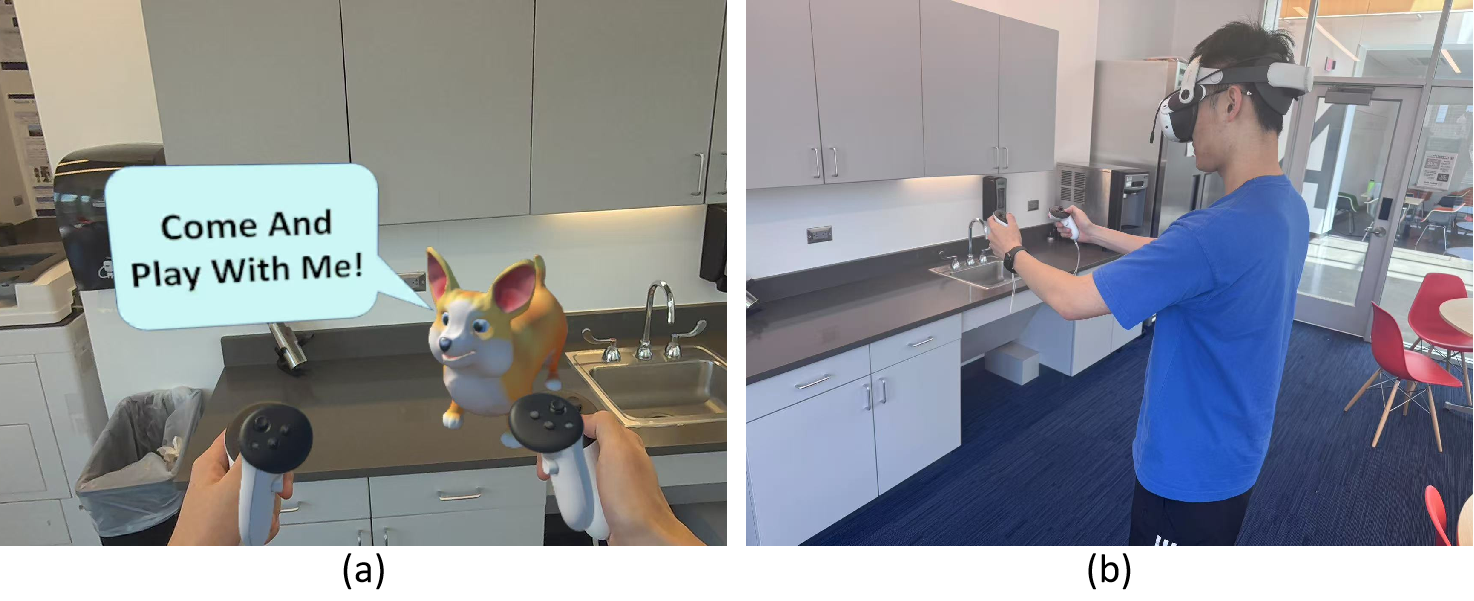}
    \caption{(a) First-person view from the AR headset: the virtual object (e.g. a dog and textual cues) is interactive and can be toggled on or off using the ``Button B" on the right controller, while ``Button A" starts or stops video recording. (b) Third-person view showing the data collector wearing the Meta Quest 3 headset and interacting with the AR environment using handheld controllers.}
    \vspace{-0.2in}
\end{figure}

\section{Video Baseline Implementation Details}

We provide the full implementation details of the supervised video baselines used in our experiments: the 3D--ResNet model~\cite{hara2017learningspatiotemporalfeatures3d} and the transformer-based ViViT model~\cite{arnab2021vivitvideovisiontransformer}. Both baselines follow standard practice in video classification and are trained under consistent data splits, preprocessing steps, and evaluation metrics. The descriptions below summarize the implementation in our released training code.%

\subsection{Dataset Splits and Sampling}

All video models use the same dataset partitions:
\begin{itemize}
    \item Train / Validation / Test: 60\% / 20\% / 20\%, stratified by attack type and scene category.
    \item One clip is extracted per video for all splits, matching the single-view evaluation protocol.
\end{itemize}

To mitigate class imbalance during training, we adopt a class-balanced sampling strategy.  
Let $c_i$ be the number of training videos for class $i$; class weights are:
\begin{equation}
    w_i = \frac{1}{c_i + 10^{-6}}, \qquad
p_i = \frac{w_i}{\sum_j w_j},
\end{equation}

and training videos are sampled according to $p_i$.  
This sampling procedure matches the implementation in our 3D--ResNet code.%

\subsection{Clip Construction and Preprocessing}

For both ViViT and 3D--ResNet, AR videos are processed into short clips using a uniform temporal sampling pipeline:
\begin{itemize}
    \item 20-frame clips:
    
    obtained using\texttt{UniformTemporalSubsample(20)}.
    \item Spatial resizing: frames are resized to 400 $\times$ 400.
    \item Pixel normalization: mean [0.45,0.45, 0.45], std [0.225, 0.225, 0.225].
\end{itemize}

Frames are arranged as tensors of shape $(T, C, H, W)$ with $T=20$.  
No additional augmentations (cropping, flipping, jittering) are applied, avoiding synthetic distortions that could alter AR semantics.

\subsection{3D--ResNet Baseline}

Our 3D CNN baseline is implemented using the \texttt{r3d\_18} architecture from \texttt{torchvision} (18-layer 3D--ResNet).%
We detail the full setup below.%

\paragraph{Architecture.}
\begin{itemize}
    \item Model: R3D-18 (3D ResNet).
    \item Pretraining: Kinetics-400 weights from Torchvision.
    \item Final classifier layer replaced with a 5-way linear head.
\end{itemize}

\paragraph{Loss Function.}
We employ a Focal Loss with $\gamma = 2.0$:
\begin{equation}
    \mathcal{L}_{\mathrm{focal}}
= (1 - \exp(-\mathrm{CE}(\mathbf{x}, y)))^\gamma \,
  \mathrm{CE}(\mathbf{x}, y),
\end{equation}

which improves stability under severe class imbalance.

\paragraph{Optimization.}
\begin{itemize}
    \item Optimizer: AdamW
    \item Learning rate: 5$\times$10$^{-5}$
    \item Weight decay: 5$\times$10$^{-3}$
    \item Scheduler: StepLR (step=10, $\gamma$=0.1)
    \item Batch size: 20
    \item Epochs: 20
\end{itemize}

\paragraph{Precision Training.}
Mixed-precision (FP16) training is performed using HuggingFace \texttt{Accelerate} with \texttt{autocast()} enabled, matching our code.%
All layers are trainable; no freezing is applied.

\subsection{ViViT Baseline}

We adopt the ViViT-B/16x2 (Factorised Encoder) model as our transformer baseline:
\begin{itemize}
    \item Patch size: 16$\times$16
    \item Hidden dimension: 768
    \item Transformer layers: 12
    \item Heads: 12
    \item Pretraining: Kinetics-400
\end{itemize}

The model receives the same 20-frame clips and performs temporal factorization before spatial tokenization as in the original ViViT design.

\paragraph{Training.}
We train ViViT for 20 epochs using:
\begin{itemize}
    \item Batch size: 16
    \item AdamW optimizer (1e$^{-4}$ learning rate, 5e$^{-3}$ weight decay)
    \item Cosine learning-rate decay
    \item Warmup steps: 10\% total steps
    \item Mixed precision (FP16) using \texttt{Accelerate}
\end{itemize}

\subsection{Evaluation Metrics}

To ensure consistency with the metrics defined in the main paper, we evaluate both video models using the following measures:

\paragraph{Class-Specific Accuracy.}
For each class $k$:
\[
\text{Acc}_k
= 
\frac{\mathrm{TP}_k}
     {\mathrm{TP}_k + \mathrm{FN}_k + \mathrm{FP}_k}.
\]

\paragraph{Precision, Recall, F1.}
Computed using scikit-learn with \texttt{zero\_division=0}.

\paragraph{Confusion Matrix.}
Collected over the test split and used to compute TP/FP/FN counts for each class.

\subsection{Fairness and Consistency}

To ensure a fair comparison with CADAR and VLM baselines:
\begin{itemize}
    \item All models receive the same raw input videos.
    \item No prompts, symbolic structures, or additional annotations are provided to video baselines.
    \item Clip preprocessing (20 frames, resizing, normalization) is identical across all supervised baselines.
\end{itemize}

Overall, these settings yield strong and transparent supervised video baselines, allowing us to isolate the contributions of CADAR's neurosymbolic reasoning.

\subsection{Model Architectures and Training}

For our supervised-learning baselines—3D ResNet and ViViT. Training hyper-parameters are shown below. We conducted all training and evaluation using Google Colab with NVIDIA A100 GPUs (40GB memory). The experiments were run in a Linux environment via Colab's remote infrastructure, conducted using PyTorch 2.6.0 with Python 3.11.

\begin{table}[h]
\small
\renewcommand{\arraystretch}{1.3}
\resizebox{\linewidth}{!}{%
\begin{tabular}{@{}l|c c c|c c c@{}}
\toprule
 & \multicolumn{3}{c|}{\textbf{3D ResNet}} & \multicolumn{3}{c}{\textbf{ViViT}} \\
\midrule
 & \textbf{Value} & \textbf{Range} & \textbf{\# Tried} & \textbf{Value} & \textbf{Range} & \textbf{\# Tried} \\
initial/end lr   & 5e-5   & [5e-5, 5e-3]     & 5 & 1e-4   & [5e-5, 5e-3]     & 4 \\
weight decay     & 5e-3   & [1e-4, 1e-1]     & 5 & 5e-3   & [1e-3, 1e-1]     & 4 \\
batch size       & 20     & [8–20]           & 3 & 16     & [4–16]           & 2 \\
epochs           & 20     & [5–50]           & 6 & 20     & [5–50]           & 4 \\
scheduler        & StepLR & --              & -- & Cosine & --              & -- \\
optimizer        & AdamW  & --              & -- & AdamW  & --              & -- \\
\bottomrule
\end{tabular}
}
\caption{Training hyper-parameters, tested ranges, and number of values tried for ViViT and 3D ResNet.}
\end{table}

We use ViViT-B/16x2 (pretrained on Kinetics-400) as the backbone for our ViViT model, and a 3D ResNet-18 for the ResNet baseline. Both models are trained on the full training set using inverse-frequency sampling to balance class distributions. The classification task involves five classes: the four attack types, along with an additional No Attack category. All input videos are resized to 400×400 pixels and normalized using a mean of [0.45, 0.45, 0.45] and a standard deviation of [0.225, 0.225, 0.225]. Each input consists of 20 uniformly sampled frames from a video clip. We use Focal Loss with a gamma of 2.0 as the training objective. Mixed precision (FP16) training is enabled via HuggingFace Accelerate for both models. ViViT is trained with a cosine learning rate scheduler, while the 3D ResNet uses a StepLR scheduler (step size = 10, gamma = 0.1). Final hyperparameter configurations for both models were selected based on the settings that yielded the highest validation accuracy.

\section{Vision–Language Model (VLM) Baseline Implementation Details}

We evaluate three state-of-the-art multimodal VLMs as baselines: GPT-5-mini (OpenAI), Gemini-2.5-flash (Google), and Grok-4-fast (xAI). All three models support multi-image inputs and natural language reasoning, enabling them to perform frame-level attack classification without any training or fine-tuning. This section details our prompt design, image preprocessing strategy, and inference procedure, following our released implementation.%

\subsection{Input Frames and Preprocessing}

Each AR video is converted into multiple reference and attack frames using the procedure in our VLM baseline code.%
We extract:
\begin{itemize}
    \item \textbf{N reference frames} uniformly from the first 30--40\% of the video, and
    \item \textbf{M attack frames} uniformly from the last 25\%.
\end{itemize}

Images are resized to a maximum resolution of $800\times800$ while preserving aspect ratio.  
For GPT-5 and Grok, images are encoded as JPEG base64 strings.  
Gemini receives PIL image objects directly, as required by the \texttt{generate\_content} interface.

\subsection{Unified Prompt for All VLM Baselines}

To ensure fair comparison across models, all VLMs use the exact same prompt (text only; no model-specific tuning).  
We introduce the four base attack types used in CADAR---text modification, visual modification, obstruction, and injection---together with detailed natural-language definitions.  
The full prompt is shown below.

\begin{quote}
You are an AR video attack analyst. Your job is to identify the presence of 4 different kinds of AR attacks and non-attacks in the video with the scene context. The definitions for each attack are as follows:

A text modification attack manipulates textual information embedded in the environment, thereby changing the semantic message conveyed by the scene. For example, replacing a ``NO PARKING FIRE LANE'' sign with ``FREE PARKING FIRE LANE'' directly reverses the intended directive and can create safety hazards. Such attacks are consequential when text provides primary contextual or navigational cues.

A visual modification attack alters an object's appearance---its color, shape, or iconography---or repositions it implausibly. Examples include transforming a pedestrian-crossing sign into a bicycle-crossing sign, switching a green traffic signal to red, or relocating a stop sign to the center of an intersection. These manipulations can be quantified by analysing pixel-level discrepancies and deviations from expected spatial distributions.

An obstruction attack hides or deletes critical objects, thereby eliminating essential information and distorting scene semantics. For example, erasing an emergency-exit sign in a fire evacuation scenario removes a safety-critical cue and disrupts the relational structure of the environment.

An Injection attack introduces new, contextually inconsistent, or misleading objects that distort scene semantics. A fake detour sign placed on a clear street, or fictitious warning icons inserted into an AR overlay, may mislead human users and perception systems. Detection relies on identifying unexpected objects that conflict with the scene’s established structure and context.

A non-attack refers to a video in which the AR content remains semantically consistent with the physical environment and does not alter the user’s understanding of the scene.

The following input will contain video frames and context. Please output the class. Note that there may be more than one kind of attack present in the video, or no attack at all. Along with the attack, mention the object the attack was performed on, or briefly describe the attack in the context of the video.
\end{quote}

This prompt is prepended to the multimodal input for all VLMs.

\subsection{Inference Procedure}

Each VLM receives:
\begin{itemize}
    \item the textual prompt above,
    \item scene context,
    \item $N=20$ reference images, and
    \item $M=5$ attack image at a time.
\end{itemize}

The models are called with their native multimodal APIs:
\begin{itemize}
    \item GPT-5-mini: OpenAI Chat Completions (image URLs encoded as base64).
    \item Gemini-2.5-flash: \texttt{generate\_content} with a mixed list of text + images.
    \item Grok-4-fast: xAI Chat Completions (OpenAI-compatible JSON format with image URLs).
\end{itemize}

Each response is required to be valid JSON, following the schema below:
\begin{verbatim}
[
  {
    "object": "<object>",
    "attack_type": "<classification>",
    "notes": "<short explanation>"
  }
]
\end{verbatim}

We apply a lightweight parser that extracts JSON blocks from the raw VLM outputs. Each model produces one prediction per attack frame.   
This mirrors video-level behavior and reduces variance inherent in single-frame VLM outputs.

\subsection{Fairness Measures}

To ensure a strict apples-to-apples comparison:
\begin{itemize}
    \item All VLMs use the same prompt with context, the same set of reference frames, and the same attack frames. 
    \item No model-specific prompt engineering, priming, or handcrafted reasoning chains are used.
    \item All text is fed to each VLM exactly once per attack frame; no iterative refinement or self-consistency sampling is applied.
\end{itemize}

Because VLMs are not adapted to CADAR’s symbolic representation or temporal structure, these baselines reflect the true performance of general-purpose open-world vision–language models on cognitive attack detection.